\documentclass[runningheads]{llncs}

\usepackage{eccv}

\usepackage{eccvabbrv}

\usepackage{graphicx}
\usepackage{booktabs}
\usepackage{marvosym}
\usepackage[accsupp]{axessibility}  

\usepackage[dvipsnames]{xcolor}

\usepackage{booktabs}
\usepackage{amsmath}
\usepackage{epsfig}
\usepackage{multirow}
\usepackage{ dsfont }
\usepackage{ cite }
\usepackage{float}
\usepackage{color, colortbl, xcolor}
\usepackage{soul}
\usepackage{float}
\usepackage{url}
\usepackage{afterpage}
\usepackage{rotate}
\usepackage{makecell}
\usepackage{multirow}
\usepackage{algorithm}
\usepackage{algorithmicx}
\usepackage{algpseudocode}
\usepackage{mathtools}
\usepackage{etoolbox}
\usepackage{lipsum}
\usepackage{bbm}
\usepackage{listings} 
\usepackage{tabularx}
\usepackage{amssymb}
\usepackage{amsbsy}
\usepackage{pifont} 
\usepackage{tikz}
\usetikzlibrary{shapes.geometric}
\usepackage[normalem]{ulem}

\usepackage{array}
\usepackage{transparent}

\definecolor{2nd}{gray}{0.8}

\newcolumntype{g}{>{\columncolor{Gray}}c}

\newcolumntype{^}{>{\currentrowstyle}}

\algdef{SE}[SUBALG]{Indent}{EndIndent}{}{\algorithmicend\ }%
\algtext*{Indent}
\algtext*{EndIndent}

\newcolumntype{Y}{>{\centering\arraybackslash}X}
\newcolumntype{P}[1]{>{\centering\arraybackslash}p{#1}}

\algnewcommand{\nComment}[1]{\Statex \Comment{#1}}
\definecolor{LightCyan}{rgb}{0.88,1,1}

\definecolor{mypurple}{RGB}{153, 0, 153}
\definecolor{mygray}{RGB}{128, 128, 128}
\definecolor{mygreen}{RGB}{0, 153, 0}

\definecolor{mycyan}{RGB}{64, 128, 128}
\definecolor{mypink}{RGB}{255, 182, 193}
\definecolor{myred}{RGB}{165,42,42}
\definecolor{myyellow}{RGB}{255, 191, 0}

\definecolor{tab_red}{rgb}{0.71, 0.11, 0.0}
\definecolor{tab_green}{rgb}{0.11, 0.71, 0.0}

\definecolor{car_yellow}{RGB}{255,255,153}
\definecolor{dog_purple}{RGB}{204,153,255}
\definecolor{bg_white}{RGB}{255,255,255}

\usetikzlibrary{shadows}
\usetikzlibrary{plotmarks}
\usetikzlibrary{shapes}

\newcommand{\thickhline}{%
	\noalign {\ifnum 0=`}\fi \hrule height 1pt
	\futurelet \reserved@a \@xhline
}
\makeatletter
\global\let\oriCT@@do@color\CT@@do@color

\usepackage[pagebackref,breaklinks,colorlinks]{hyperref}

\usepackage{orcidlink}

\begin{document}

\title{Learning to Complement and to Defer to Multiple Users} 

\titlerunning{Learning to Complement and to Defer to Multiple Users}

\author{Zheng Zhang\inst{1}\orcidlink{0000-0002-1805-7705} \and
Wenjie Ai\inst{1}\orcidlink{0009-0003-4387-8745} \and
Kevin Wells\inst{1}\orcidlink{0000-0002-4658-8060} \and
David Rosewarne\inst{1,2} \and
Thanh-Toan Do\inst{3}\orcidlink{0000-0002-6249-0848} \and
Gustavo Carneiro\inst{1}\textsuperscript{\Letter}\orcidlink{0000-0002-5571-6220}}

\authorrunning{Z.~Zhang et al.}

\institute{Centre for Vision, Speech and Signal Processing, University of Surrey, UK \and
Royal Wolverhampton Hospitals NHS Trust, UK
\and
Department of Data Science and AI, Monash University, Australia\\}

\maketitle

\setcounter{footnote}{0}
\begin{abstract}

With the development of Human-AI Collaboration in Classification (HAI-CC), integrating users and AI predictions becomes challenging due to the complex decision-making process.
This process has three options: 1) AI autonomously classifies, 2) learning to complement, where AI collaborates with users, and 3) learning to defer, where AI defers to users. 
Despite their interconnected nature, these options have been studied in isolation rather than as components of a unified system.
  In this paper, we address this weakness with  the novel HAI-CC methodology, called \underline{Le}arning to \underline{Co}mplement and to \underline{D}efer to Multiple \underline{U}sers (LECODU).
  LECODU not only combines \textit{learning to complement} and \textit{learning to defer} strategies, but it also incorporates an estimation of the optimal number of users to engage in the decision process.
  The training of LECODU maximises classification accuracy and minimises collaboration costs associated with user involvement. 
  Comprehensive evaluations across real-world and synthesized datasets demonstrate LECODU's superior performance compared to state-of-the-art HAI-CC methods. Remarkably, even when relying on unreliable users with high rates of label noise, LECODU exhibits significant improvement over both human decision-makers alone and AI alone\footnote{Supported by the Engineering and Physical Sciences Research Council (EPSRC) through grant EP/Y018036/1.}. Code is available at \href{https://github.com/zhengzhang37/LECODU.git}{https://github.com/zhengzhang37/LECODU.git}
  .
\end{abstract}

\section{Introduction}
\label{sec:intro}

Human-AI collaborative classification (HAI-CC) optimally combines human and AI strengths in training and testing, offering superiority over standalone AI models and manual decision processes~\cite{dafoe2021-cooperative,steyvers2022bayesian}.
For instance, HAI-CC tends to be more accurate, interpretable, and engaging than standalone AI models~\cite{dafoe2021-cooperative,steyvers2022bayesian}.
Compared with manual processes, HAI-CC can enhance efficiency and decision consistency, and mitigate human errors~\cite{dafoe2021-cooperative,steyvers2022bayesian}.
Thus, HAI-CC is increasingly recognized as a valuable classification strategy, particularly for complex problems in high-stake real-world scenarios, such as breast cancer classification from mammograms~\cite{halling2020optimam}, risk assessment endeavors~\cite{green2019disparate},
and the identification of incorrect or misleading information generated by large language models~\cite{wei2022emergent,bubeck2023sparks}.

HAI-CC approaches~\cite{dafoe2021-cooperative} aim to maximise accuracy and minimise the human-AI collaboration costs through learning to defer and learning to complement techniques. 
In learning to defer~\cite{mozannar2020consistent}, HAI-CC determines when to classify with the AI model alone or defer to users, while learning to complement~\cite{complement_wilder} combines AI and user classifications. 
In a single expert HAI-CC (SEHAI-CC) setting, only one user is included~\cite{narasimhan2022post,raghu2019algorithmic,okati2021differentiable,mozannar2020consistent,verma2022calibrated,whoshould_mozannar23,complement_wilder,charusaie2022sample}, while multiple experts HAI-CC (MEHAI-CC) methods explore strategies to complement with~\cite{ijcai2022-344} or defer to multiple experts~\cite{verma2022calibration,mao2023two,multil2d,keswani2021towards}.
Despite the interconnected nature of learning to defer and learning to complement, prior HAI-CC methods have mainly studied them in isolation. 
This oversight represents a missed opportunity to adopt a more comprehensive approach that could improve HAI-CC efficacy.
Additionally, most HAI-CC methods assume the presence of clean labels in the training set, which may be impractical, especially in multi-rater training sets where clean labels may be unavailable~\cite{zhu2021hard} or impossible to recover. Among HAI-CC techniques, LECOMH~\cite{zhang2023learning} stands out as an exception that does not depend on clean labels, relying on noisy-label learning techniques~\cite{song2022learning} to handle multiple noisy labels during training. Nevertheless, LECOMH is a learning to complement approach that does not consider the learning to defer option.

\begin{figure}[t!]
	\centering
        \scalebox{0.6}{
	\includegraphics[width=\linewidth]{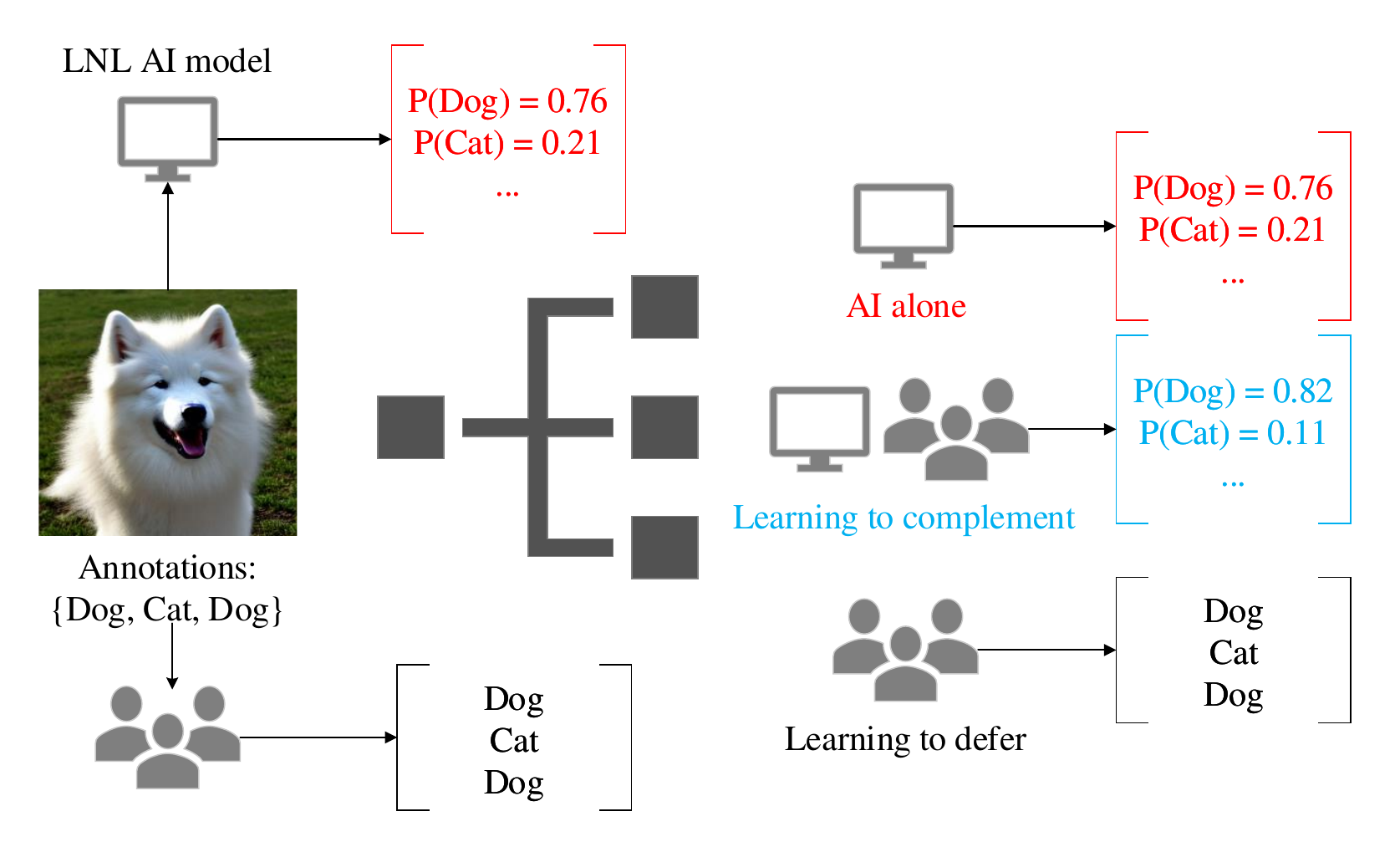}}
\caption{LECODU, our proposed Human-AI Collaborative Classification (HAI-CC) methodology, integrates standalone AI classification with the concepts of learning to defer~\cite{mozannar2020consistent} and learning to complement~\cite{complement_wilder}. Our training process explores learning with noisy-label (LNL) techniques to leverage a multi-rater training set (where clean labels are not available) to maximise the HAI-CC accuracy and minimise the collaboration costs represented by the number of users involved in collaborative classification.}
\label{fig:framework}
\end{figure}


In this paper, we propose a novel HAI-CC approach called \underline{Le}arning to \underline{Co}mplement and to \underline{D}efer to Multiple \underline{U}sers (LECODU). 
LECODU (Fig.~\ref{fig:framework}) integrates learning to defer and learning to complement to multiple users, leveraging training sets containing multiple noisy-label annotations.
More specifically, LECODU is designed to make three decisions: 1) when to collaborate with experts, 2) when to defer to experts, and 3) how many experts should be engaged in the decision process. 
These three decisions target the highest classification accuracy at the minimum collaboration cost, where cost is measured by the number of experts engaged in the decision process.
Our main contributions are:
\begin{itemize}
    \item The novel LECODU model featuring a selection module and a collaboration module that combines learning-to-defer and learning-to-complement strategies; and
    \item A new training algorithm that leverages a training set containing multiple noisy labels per image to minimise the costs and maximise the accuracy of LECODU.
\end{itemize}
We assess LECODU's effectiveness against state-of-the-art (SOTA) HAI-CC methods~\cite{whoshould_mozannar23,ijcai2022-344,multil2d, zhang2023learning} using real-world and synthesized multi-rater noisy label benchmarks (CIFAR-10N~\cite{wei2021learning}, CIFAR-10H~\cite{peterson2019human}, multi-rater CIFAR10-IDN~\cite{xia2021sample}, and Chaoyang~\cite{zhu2021hard}). 
Across all benchmarks, LECODU consistently outperforms the competition, showing higher accuracy for equivalent collaboration costs, as measured by the number of experts engaged in classification. Even in scenarios with medium and high noise rates where user annotations are unreliable, LECODU exhibits remarkable improvement over both human decision-makers and AI alone.

\section{Related work}

\noindent
\textbf{Human-AI Collaborative Classification (HAI-CC)} explores the strengths of expert users and AI by offering improved accuracy, interpretability, and ethical considerations compared to standalone AI models, while also providing increased efficiency, data-driven insights, and consistency in decision-making compared to users without AI assistance~\cite{lu2021human, yin2019understanding,chiou2023trusting,shin2021effects, weitz2019you}. 
HAI-CC is motivated by recent studies~\cite{rosenfeld2018totally, serre2019deep,kamar2012combining} which suggest that previous AI classification approaches have predominantly focused on optimising AI systems in isolation, neglecting the influence of human-AI collaborative classification.
To address this gap, many methods have been proposed to harmonize and enhance the collaboration between humans and AI~\cite{bansal2021most, agarwal2023combining, vodrahalli2022humans, humanIntheloop, complement_wilder,pradier2021preferential}.

A prevalent HAI-CC strategy is Learning to Defer (L2D), which is optimised to decide if the expert user or the AI will make the classification. Originated from the learning to reject strategy~\cite{cortes2016learning}, which considered the impact of other agents in the decision-making process~\cite{madras2018predict}, L2D was addressed by studies on deferral and abstention through optimising different surrogate loss functions of the classification loss~\cite{narasimhan2022post,raghu2019algorithmic,okati2021differentiable,mozannar2020consistent,verma2022calibrated,whoshould_mozannar23,charusaie2022sample,cao2024defense}. 
Mozannar~\etal~\cite{mozannar2020consistent} proposed a cost-sensitive softmax cross-entropy consistent surrogate loss, while Verma~\etal~\cite{verma2022calibrated} suggested a one-vs-all classifier as another consistent surrogate to mitigate underfitting issues. 
Disagreement on disagreements (DoD)~\cite{charusaie2022sample} provided an active learning algorithm that is able to train a classifier-rejector pair by minimally querying the human on selected points.
Furthermore, Mozannar~\etal~\cite{whoshould_mozannar23} provided a realizable-consistent surrogate loss function to address learning with deferral using halfspaces.
A common limitation across these studies is their focus on single-expert scenarios, overlooking the complexities of multi-user collaboration within human-AI teams.
Recently, several works have focused on learning to defer with multiple users~\cite{verma2022calibration,mao2023two,ijcai2022-344,multil2d,keswani2021towards,babbar2022utility}.
Hemmer~\etal~\cite{ijcai2022-344} 
introduced a model with ensemble prediction combining AI and human predictions, but collaboration cost optimization is lacking. 
On the other hand, Mozannar~\etal~\cite{multil2d} proposed an L2D method capable of deferring to one of multiple users without combining AI and human predictions. Mao~\etal~\cite{mao2023two} propose a two-stage $\mathcal{H}$-consistent surrogate loss for learning to defer to multiple experts. 

Another important HAI-CC strategy, learning to complement (L2C)~\cite{complement_wilder}, has been designed to maximise the expected utility of the human-AI decision. 
Steyvers~\etal~\cite{steyvers2022bayesian} proposed a Bayesian framework for modeling human-AI complementarity, while Kerrigan~\etal~\cite{kerrigan2021combining} proposed to combine human and model predictions via confusion matrices 
and model calibration. 
Zhang~\etal~\cite{zhang2023learning} proposed LECOMH, integrating learning to complement with learning with noisy label (LNL) and multi-rater learning (MRL). 
Bansal~\etal~\cite{bansal2021most} focus on optimising the expected utility of decision-making for human-AI teams, diverging from traditional model optimization centered on  accuracy.
Liu~\etal~\cite{liu2023humans} leverage perceptual differences through post-hoc teaming, showing that human-machine collaboration can be more accurate than machine-machine collaboration.

We reiterate that L2D and L2C methods have similar goals, but have been developed in isolation, even though they should be considered parts of a unified system. This is a gap we are addressing in this paper with LECODU.

\noindent
\textbf{Learning with Noisy Labels (LNL).}
Except for LECOMH~\cite{zhang2023learning}, previous HAI-CC methods do not rely on any technique to handle label noise in the training set, which is critical for dealing with real-world data that commonly contains multiple noisy labels per training sample. 
In this subsection, we provide a short review of LNL methods that can be used in the HAI-CC context.
LNL approaches have explored many techniques, including: robust loss functions~\cite{zhang2018generalized,ghosh2017robust}, 
co-teaching~\cite{MentorNet, han2018co},
label cleaning \cite{yuan2018iterative, jaehwan2019photometric}, 
semi-supervised learning (SSL)~\cite{ li2020dividemix, ortego2021multi}, 
iterative label correction~\cite{label_clean,arazo2019unsupervised}, 
meta-learning ~\cite{L2W,Distill_noise,FSR,Famus}, and 
graphical models~\cite{garg2023instance}.
Existing LNL SOTA methods predominantly rely on SSL techniques. 
For instance, DivideMix~\cite{li2020dividemix} integrates MixMatch~\cite{berthelot2019mixmatch} and Co-teaching~\cite{han2018co} to harness the SSL potential.
Adhering to this paradigm, several LNL studies employ MixUp~\cite{zhang2017mixup} within SSL~\cite{cordeiro2023longremix, sachdeva2023scanmix, zhu2021hard}. 
Furthermore, InstanceGM~\cite{garg2023instance} introduces graphical models to work together with SSL methods, while Promix~\cite{wang2022promix} optimises the utility of clean samples through a matched high-confidence selection technique.
However, LNL is inherently an ill-posed problem that needs constraints to become identifiable. One such constraint is the presence of multiple noisy labels per training sample~\cite{liu2023identifiability}, which is naturally present in multiple expert HAI-CC (MEHAI-CC). Therefore, investigating LNL techniques in MEHAI-CC appears to be a promising exploration  avenue.  Such techniques are known as multi-rater learning (MRL) and are described in more detail next.

\noindent
\textbf{Multi-rater Learning (MRL)} aims to train a classifier with noisy crowdsourced labels from multiple annotators, where the key is how to extract the ``clean'' label from the imperfect crowdsourced labels. Previous methods focus on majority voting~\cite{zhou2012ensemble} and expectation-maximization (EM) algorithm~\cite{rodrigues2014gaussian,rodrigues2017learning,dawid1979maximum,whitehill2009whose, raykar2010learning}.
CrowdLayer~\cite{rodrigues2018deep} trains an end-to-end deep neural network (DNN) with crowdsourced labels through a crowd layer that models the annotator-specific transition matrix. 
SpeeLFC~\cite{chen2021structured} proposes a probabilistic model that learns an interpretable transition matrix for each annotator. 
WDN~\cite{guan2018said} trains the classifier with multiple output layers and learns combination weights for aggregating the outputs.
Recently, UnionNet-B~\cite{wei2022deep} is proposed to take the labelling information provided by all annotators as a union and coordinate multiple annotators, while CROWDLAB~\cite{crowdlab} achieves SOTA MRL results by relying on multiple noisy-label samples and predictions with an LNL-trained model. Even though MRL is a promising research direction to leverage multiple noisy labels per training sample, it disregards the idea of human-AI collaboration for training and testing, which is a point explored by HAI-CC methods, e.g., LECODU.

\section{Method}
\vspace{-5pt}
\begin{figure}[t!]
	\centering
        \scalebox{0.98}{
	\includegraphics[width=\linewidth]{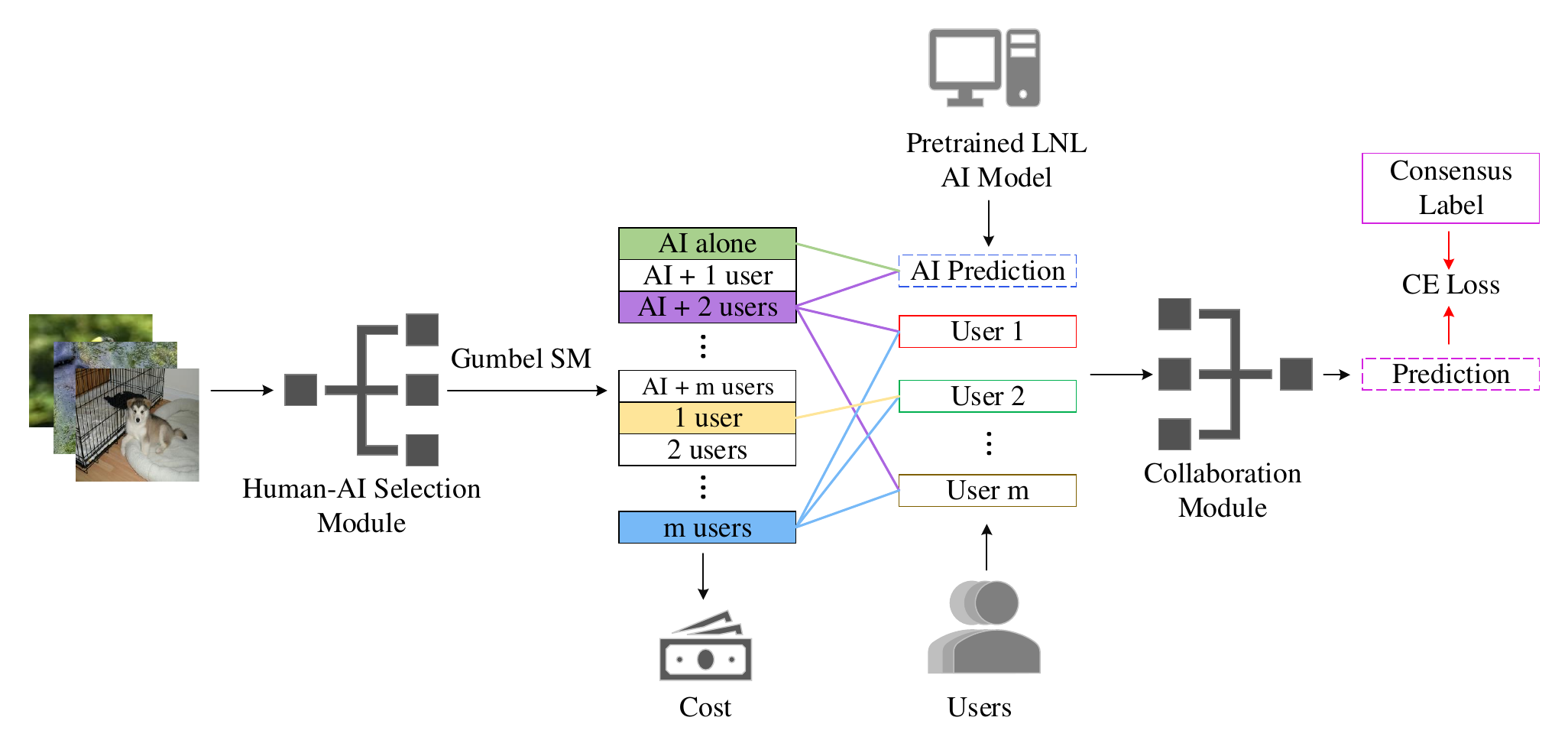}}
	\caption{
 LECODU contains a Human-AI Selection Module and a Collaboration Module. The Human-AI Selection Module aims to decide whether we use pre-trained LNL AI model~\cite{wang2022promix,zhu2021hard,garg2023instance} alone, complement the LNL AI model's prediction with $\{1,...,M\}$ users, or defer the decision to an ensemble of $\{1,...,M\}$ users, where the system cost is based on the number of users in this collaboration. The Collaboration Module uses the collaborative option selected by the Human-AI Selection Module (AI alone, AI + 1 user, ..., AI + $M$ users, 1 user, ..., $M$ users) to make a final classification.}
	\label{fig:architecture}
\end{figure}
Let $\mathcal{D}=\{x_i,\mathcal{M}_i\}^{|\mathcal{D}|}_{i=1}$ be the noisy-label multi-rater training set, where $x_i\in\mathcal{X}\subset\mathbb{R}^{H\times W\times R}$ denotes an input image of size $H\times W$ and $R$ channels, 
and $\mathcal{M}_i=\{ m_{i,j}\}_{j=1}^{M}$ represents the $M$ experts' noisy annotations for image $i$, with $m_{i,j} \in \mathcal{Y} \subset \{0,1\}^{|\mathcal{Y}|}$ being a one-hot label. A SOTA LNL AI model~\cite{wang2022promix,garg2023instance,zhu2021hard} $f_\theta:\mathcal{X} \to \Delta^{|\mathcal{{Y}}|-1}$ has been pre-trained with the training set (the noisy label per image $x_i$ is randomly selected as one of the experts' annotations in $\mathcal{M}_i$), where $\Delta^{|\mathcal{Y}|-1}$ denotes the $|\mathcal{Y}|$-dimensional probability simplex, and $\theta\in\Theta$ is the model parameter. The consensus label for training our LECODU is obtained via the SOTA MRL method CROWDLAB~\cite{crowdlab} that takes the training images and experts' labels  $(x,\mathcal{M}) \in \mathcal{D}$, together with the AI classifier's predictions $\hat{y} = f_{\theta}(\mathbf{x})$ for each sample in $\mathcal{D}$ to produce a consensus label $\hat{y}^c \in \mathcal{Y}$ and a quality (or confidence) score $\alpha$. 
The consensus label dataset is formed with:
\begin{equation}
  \mathcal{D}^c =\{(x_i,\hat{y}^c_i,\mathcal{M}_i)|(x_i,\mathcal{M}_i)\in\mathcal{D}, (\hat{y}^c_i,\alpha_i) = \mathsf{CrowdLab}(x_i,f_{\theta}(x_i),\mathcal{M}_i), \alpha_i>0.5\}.
  \label{eq:consensus}
\end{equation}
Our proposed LECODU is shown in \cref{fig:architecture}, which contains a Human-AI Selection Module and a Collaboration Module. 
The Human-AI Selection Module, denoted as $g_{\phi}:\mathcal{X} \to \Delta^{2M}$, is designed to predict a categorical distribution reflecting the probability distribution over several collaborative scenarios, which allows for an integration of AI prediction with human experts' predictions across these different  collaborative scenarios.
These scenarios include the standalone AI prediction (1st dimension),
the learning to complement prediction that integrates the AI with one or multiple experts (2nd dimension: AI +$1$ user, ..., M+1st dimension: AI + $M$ users), 
and the learning to defer prediction that leaves the prediction to be done collectively by one or multiple experts (M+2nd dimension: $1$ user, ..., 2M+1st dimension: $M$ users). 
The Collaboration Module $h_{\psi}:\underbrace{\Delta^{|\mathcal{Y}|-1} \times  ... \times \Delta^{|\mathcal{Y}|-1}}_{2M+1 \text{ times}} \to \Delta^{|\mathcal{Y}|-1}$ 
has the AI prediction alone as its first input, the next inputs, indexed by $2$ to $M+1$, represent the collaborative prediction integrating AI with 1 to $M$ users, and inputs $M+2$ to $2M+1$ denote the collective predictions of 1 to $M$ users.
To be unbiased to any specific expert, we randomly selected users from the pool of users during training and testing for both the learning to complement and learning to defer scenarios.

The training for the Human-AI Selection Module and the Collaboration Module relies on the following optimisation:
\begin{equation}
\scalebox{1}{$
    \begin{split}
        \phi^*, \psi^* = \arg\min_{\phi,\psi} \frac{1}{|\mathcal{D}^c|} \sum_{(x_i,\hat{y}^c_i,\mathcal{M}_i) \in \mathcal{D}^c} & \ell \left (\hat{y}^c_i,h_{\psi}\left ( \mathsf{p}\left (g_{\phi}(x_i),f_{\theta}(x_i), \mathsf{shf}(\mathcal{M}_i) \right ) \right ) \right ) \\
        & + \lambda \times \mathsf{cost}(g_{\phi}(x_i)),
    \end{split}
    $}
    \label{eq:loss_function}
\end{equation}
where $\ell(.)$ is the cross-entropy (CE) loss, $\lambda$ is a hyper-parameter that weights the cost function,
\begin{equation}
\scalebox{0.9}{$
   \mathsf{p}  \left (g_{\phi}(x),f_{\theta}(x), \mathsf{shf}(\mathcal{M}) \right ) = \begin{cases} [f_{\theta}(x),\mathbf{0}_{|\mathcal{Y}|},...,\mathbf{0}_{|\mathcal{Y}|}] & \mbox{if } \max_j g^{(j)}_{\phi}(x) = g_{\phi}^{(1)}(x) \\   [f_{\theta}(x),m_{i,1},...,\mathbf{0}_{|\mathcal{Y}|}] & \mbox{if } \max_j g^{(j)}_{\phi}(x) = g_{\phi}^{(2)}(x) \\ 
     ... & \\
   [f_{\theta}(x),m_{i,1},...,m_{i,M}] & \mbox{if } \max_j g^{(j)}_{\phi}(x) = g_{\phi}^{(M+1)}(x) \\
   [\mathbf{0}_{|\mathcal{Y}|},m_{i,1},...,\mathbf{0}_{|\mathcal{Y}|}] & \mbox{if } \max_j g^{(j)}_{\phi}(x) = g_{\phi}^{(M+2)}(x) \\
   ... & \\
   [\mathbf{0}_{|\mathcal{Y}|},m_{i,1},...,m_{i,M}] & \mbox{if } \max_j g^{(j)}_{\phi}(x) = g_{\phi}^{(2M+1)}(x) \\
   \end{cases},
   $}
\label{eq:concatenate_predictions}
\end{equation}
with $g^{(j)}_{\phi}(.)$ denoting the $j^{th}$ output from the Human-AI Selection Module and $\mathsf{shf}(\mathcal{M})$ representing a function that shuffles the users' annotations (so the training is unbiased to any specific users' annotations), and
\begin{equation}
\scalebox{0.9}{$
    \mathsf{cost}(g_{\phi}(x)) = \sum_{j=1}^{M+1}g^{(j)}_{\phi}(x) \times (j-1) + \sum_{j=M+2}^{2M+1}g^{(j)}_{\phi}(x) \times (j-M-1).
    $}
    \label{eq:cost}
\end{equation}
In Equation~\ref{eq:cost}, when the AI model provides the prediction alone, we have \\ $\max_j g^{(j)}_{\phi}(x) = g_{\phi}^{(1)}(x)$, with $\mathsf{cost}(g_{\phi}(x)) = 0$, but as $\max_j g^{(j)}_{\phi}(x) = g_{\phi}^{(K)}(x)$ for $K\in[2,M+1]$ (i.e., learning to complement with single/multiple users), we have $\mathsf{cost}(g_{\phi}(x)) = K-1$ and for $K\in[M+2,2M+1]$ (i.e., learning to defer with single/multiple users), we have $\mathsf{cost}(g_{\phi}(x)) = K - M - 1$. In other words, the cost in Equation~\ref{eq:cost} denotes cost of one unit per user's annotation.

In LECODU's training and testing, as illustrated in~\cref{fig:architecture}, the system starts with the concatenation of AI predictions produced by $f_{\theta}(.)$ with the annotations provided by users in $\mathcal{M}$. Then, the system selects the collaboration format (AI alone, AI + 1 user, ..., AI + $M$ users, 1 user, ..., $M$ users) by leveraging the predictions from the Human-AI Selection Module, as detailed in Equation~\ref{eq:concatenate_predictions}. 
The selection of a collaboration format is based on a non-differentiable term $\max_j g^{(j)}_{\phi}(x)$. Therefore, we re-parameterize $\max_j g^{(j)}_{\phi}(x)$ via Gumbel Softmax~\cite{jang2016categorical}, which is a re-parameterisation trick to provide a continuous approximation to the categorical distribution on the simplex $\Delta^{2M}$, enabling the computation of the parameter gradients.
Then, the Collaboration Module integrates the selected AI and user annotations to make decisions. 

\section{Experiments}
\vspace{-5pt}
\subsection{Datasets}
\noindent
\textbf{CIFAR-10 Dataset.} CIFAR-10~\cite{krizhevsky2009learning} has 50K training images and 10K testing images of size $32\times 32$, with each image belonging to one of 10 classes. To leverage \textit{real-world human annotations} on CIFAR-10, we utilize CIFAR-10N~\cite{wei2021learning} for training and CIFAR-10H~\cite{peterson2019human} for testing. 
CIFAR-10N extends the CIFAR-10 training set with three noisy labels for each image, enriching the dataset with real-world annotation variability. 
CIFAR-10H extends the CIFAR-10 test set by providing around 51 noisy labels per image.
Following LECOMH setting~\cite{zhang2023learning}, in the testing phase, a maximum of three users are randomly selected from the the pool of 51 users in CIFAR-10H to engage in collaboration.
Secondly, we conduct experiments with multi-rater CIFAR10-IDN~\cite{xia2021sample}, which harnesses \textit{synthesised annotations} with multi-rater instance-dependent label noise. The label noise rates range from 0.2 to 0.5 for both training and testing sets. Three distinct noisy labels are produced for each noise rate to enable the simulation of diverse human predictions with similar error rates.

\noindent
\textbf{Chaoyang Dataset.} 
The Chaoyang dataset comprises 6,160 colon slide patches, each with a resolution of $512\times 512$~\cite{zhu2021hard}, where each patch has \textit{three noisy labels produced by real pathologists}, with each image belonging to one of 4 classes. In the original Chaoyang dataset setup, the training set has patches with multi-rater noisy labels, while the testing set only contains patches that all users agree on a single label. 
To ensure that both training and testing sets contain multiple noisy labels and no patient data overlaps between the two sets, we follow the LECOMH data partition~\cite{zhang2023learning}, where the  new training set has 862 patches with 2 out of 3 consensual labels and 3862 patches with 3 out of 3 consensual labels. The new testing set includes 449 patches with 2 out of 3 consensual labels and 986 patches with 3 out of 3 consensual labels.

\subsection{Implementation Details}
\noindent
\textbf{Architecture.} 
All methods are implemented in Pytorch~\cite{paszke2019pytorch} and run on NVIDIA RTX A6000. 
For CIFAR-10N and CIFAR-10H, we pre-trained ProMix~\cite{wang2022promix} with two PreAct-ResNet-18 as the LNL AI model using the Rand1~\cite{wei2021learning} annotation. 
For the multi-rater CIFAR10-IDN experiments, we pre-trained 
InstanceGM~\cite{garg2023instance} with two PreAct-ResNet-18 as the LNL AI model. 
For Chaoyang, following NSHE~\cite{zhu2021hard}, two ResNet-34 are pre-trained using the label\_A annotation, and the best-trained network is selected for the LNL AI model. 
All the above models were selected because of their SOTA performance in the respective datasets.
For the Human-AI Selection Module, we utilise the same backbone of the pre-trained models. The Collaboration Module comprises a two-layer MLP with 512-dimensional hidden layers with a ReLU activation function. 
The pre-trained Promix on CIFAR-10N reaches 97.41\% accuracy on the CIFAR-10 test set.
On the multi-rater CIFAR-10 IDN, the pre-trained InstanceGM reaches test accuracy of 96.64\%, 96.52\%, 96.33\% and 95.90\% for noise rates 0.2, 0.3, 0.4 and 0.5.
The pre-trained NSHE reaches 82.44\% test accuracy on Chaoyang. 

\noindent
\textbf{Training and evaluation details.}
LECODU is trained for 200 epochs using SGD with a momentum=0.9 and a weight decay=0.0005. The batch size is 1024 for CIFAR and 96 for Chaoyang, with an initial learning rate of 0.05. The Gumbel Softmax temperature parameter of the Human-AI Selection Module is 5. 
Also, to ensure consistent data range, the AI predictions are normalised by Gumbel Softmax with temperature parameter of 0.5 before the concatenation with user annotations.
To prevent bias, the order of human annotations is randomly shuffled during both training and testing, drawn from the human label pool for each image.
Ground truth training labels are determined using CROWDLAB's consensus label. Evaluation is based on test accuracy (\%) relative to collaboration cost, measured by the number of labels used from the human label pool for the whole testing set.

\noindent
\textbf{HAI-CC Baselines.} 
Following~\cite{whoshould_mozannar23,zhang2023learning}, we compare LECODU with
single-expert L2D (SEL2D) methods, such as cross-entropy surrogate~\cite{mozannar2020consistent} (CE), one-vs-all-based surrogate~\cite{verma2022calibrated} (OvA), selective prediction that thresholds classifier confidence for the rejector~\cite{whoshould_mozannar23} (SP), the confidence method~\cite{raghu2019algorithmic} (CC), differentiable triage~\cite{okati2021differentiable} (DIFT), mixture of experts~\cite{madras2018predict}  (MoE), and Realizable Surrogate (RS)~\cite{whoshould_mozannar23}.
For training SEL2D methods, we rely on randomly sampled annotations or aggregated (majority voting) annotations to simulate a single expert from the human annotation pools, while for testing, we use randomly sampled annotations.
To compute SEL2D's collaboration cost, we sort the testing images based on their rejection scores and then adjust the threshold for annotating these testing cases by users~\cite{whoshould_mozannar23}. 
We also compare LECODU with multi-expert L2D (MEL2D) method Multi\_L2D~\cite{multil2d} and multi-expert L2C (MEL2C) methods: classifier and expert team (CET)~\cite{ijcai2022-344}, and learning to complement with multiple humans (LECOMH)~\cite{zhang2023learning}, where the number of experts is three (i.e., the maximum number of experts in our benchmarks). 
For a fair comparison, all classification backbones for the methods have the same benchmark architecture.
All SEL2D methods rely on LNL pre-training and the consensus label because it provides better results for all cases, but for CET and Multi\_L2D, we show training with (w. LNL) and without (w/o LNL) LNL pre-training (their consensus labels are produced from their different pre-training networks and 3 annotations). For LECOMH, we display the results with LNL pre-training since it provides better results for all cases. For all methods, hyper-parameters are set as previously reported in~\cite{whoshould_mozannar23,ijcai2022-344,multil2d, zhang2023learning}.

\subsection{Results}

For CIFAR-10, the total cost computation is governed by Eq.~\ref{eq:cost}, where the parameter $\lambda$ in Eq.~\ref{eq:loss_function} is adjusted during the training phase to influence LECODU's cost considerations. 
This results in a minimum cost of 0 (all testing cases predicted by AI alone) and a maximum cost of 30000 (all testing cases predicted by AI + 3 users or deferred to 3 users) for 10K test images. In contrast, for  Chaoyang dataset, the cost scale is normalized from the original span of $[0, 3\times 1435]$ to a broader range of $[0, 3 \times 10000]$ to enable an easier comparison across different datasets. Single-expert methods have a total cost in [0, 10000] as only one user per image is allowed. LECODU assesses accuracy within the cost range of [0, 10000], and for multi-expert methods, to maintain consistency in comparative analysis, if cost $>$ 10000, the accuracy plot is truncated at cost=10000. 

\begin{figure}[t!]
  \centering
    \begin{subfigure}{.32\linewidth}
    \centering
    \includegraphics[width=\linewidth]{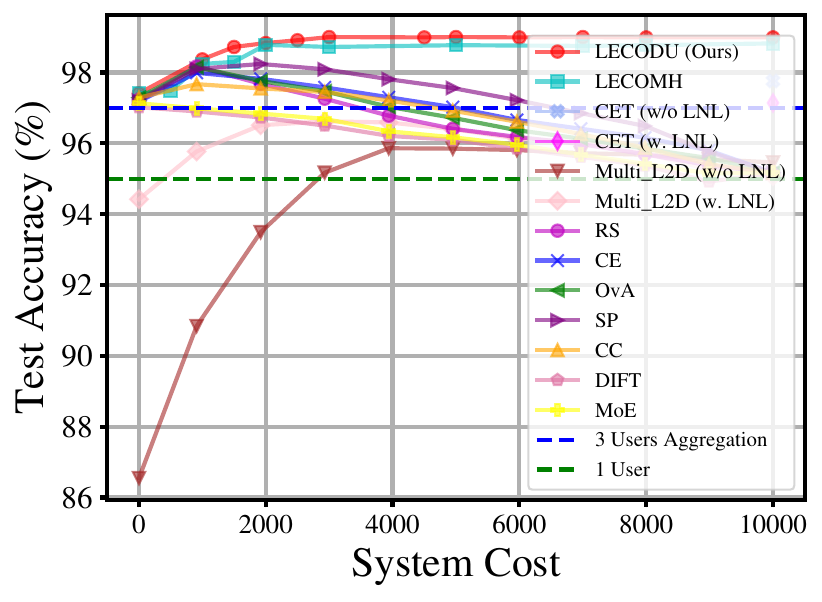}
    \caption{CIFAR-10H-aggregation.}
    \label{fig:cifarh_aggre_exp}
  \end{subfigure}
  \begin{subfigure}{.32\linewidth}
    \centering
    \includegraphics[width=\linewidth]{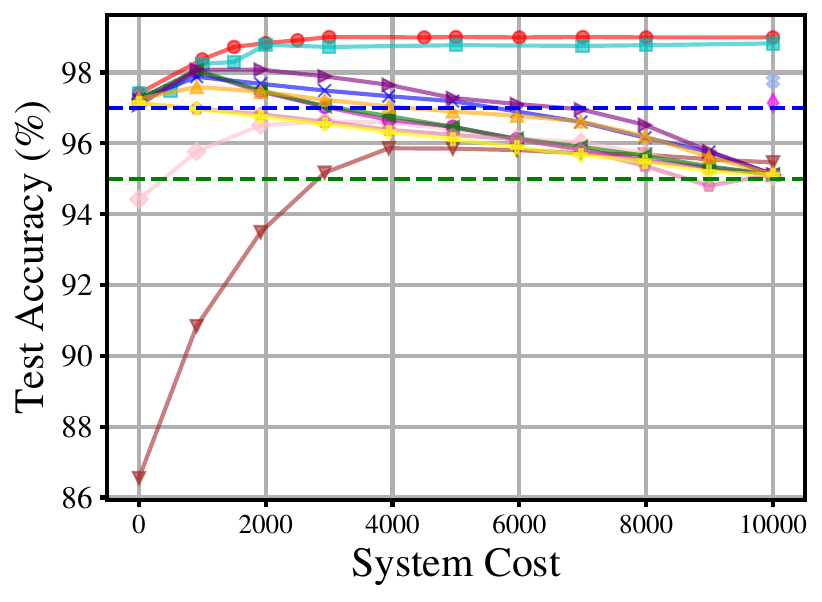}
    \caption{CIFAR-10H-random.}
    \label{fig:cifarh_random_exp}
  \end{subfigure}
  \begin{subfigure}{.32\linewidth}
    \centering
    \includegraphics[width=\linewidth]{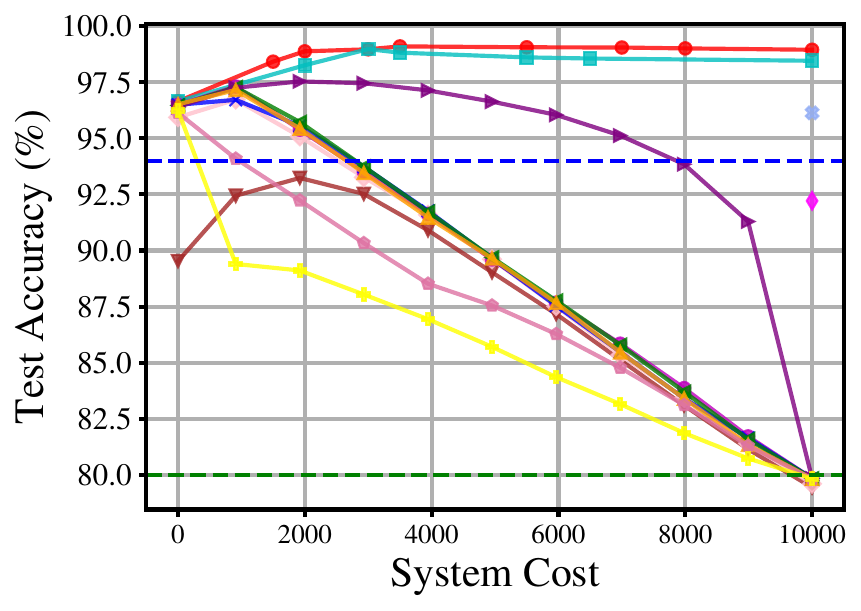}
    \caption{IDN20-aggregation.}
    \label{fig:idn20_aggre_exp}
  \end{subfigure}
  \begin{subfigure}{.32\linewidth}
    \centering
    \includegraphics[width=\linewidth]{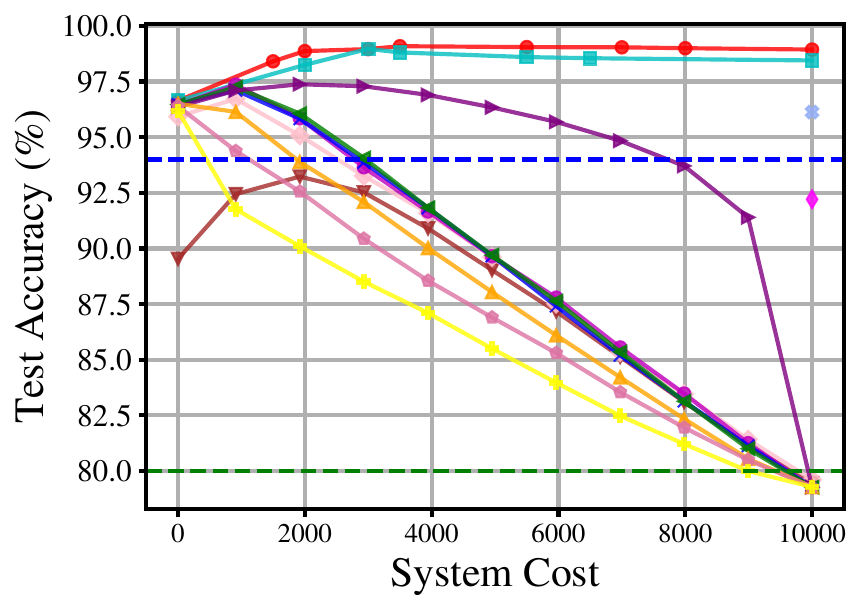}
    \caption{IDN20-random.}
    \label{fig:idn20_random_exp}
  \end{subfigure}
  \begin{subfigure}{.32\linewidth}
    \centering
    \includegraphics[width=\linewidth]{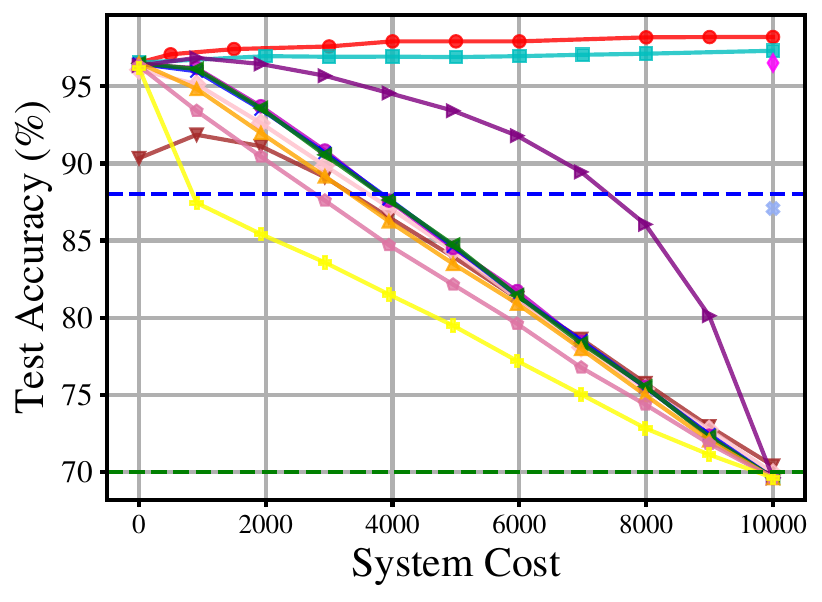}
    \caption{IDN30-aggregation.}
    \label{fig:idn30_aggre_exp}
  \end{subfigure}
  \begin{subfigure}{.32\linewidth}
    \centering
    \includegraphics[width=\linewidth]{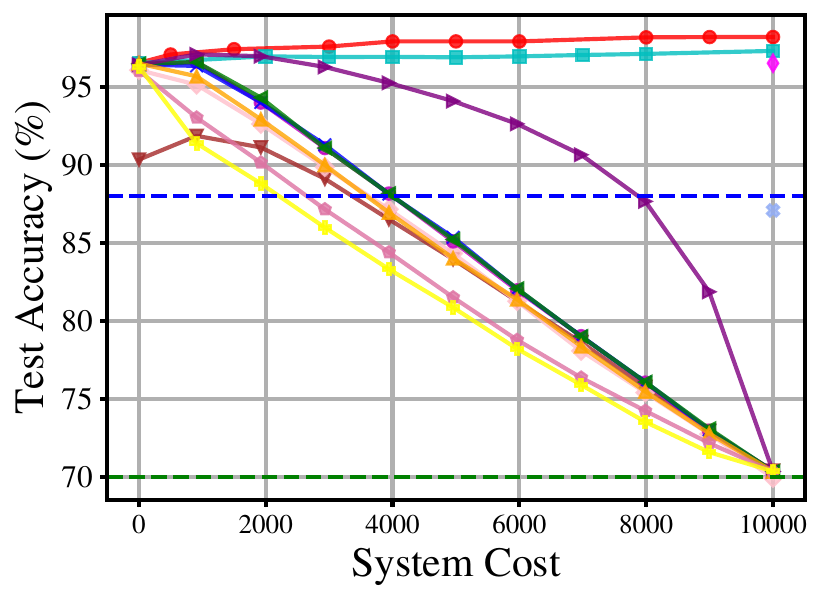}
    \caption{IDN30-random.}
    \label{fig:idn30_random_exp}
  \end{subfigure}
  \begin{subfigure}{.32\linewidth}
    \centering
    \includegraphics[width=\linewidth]{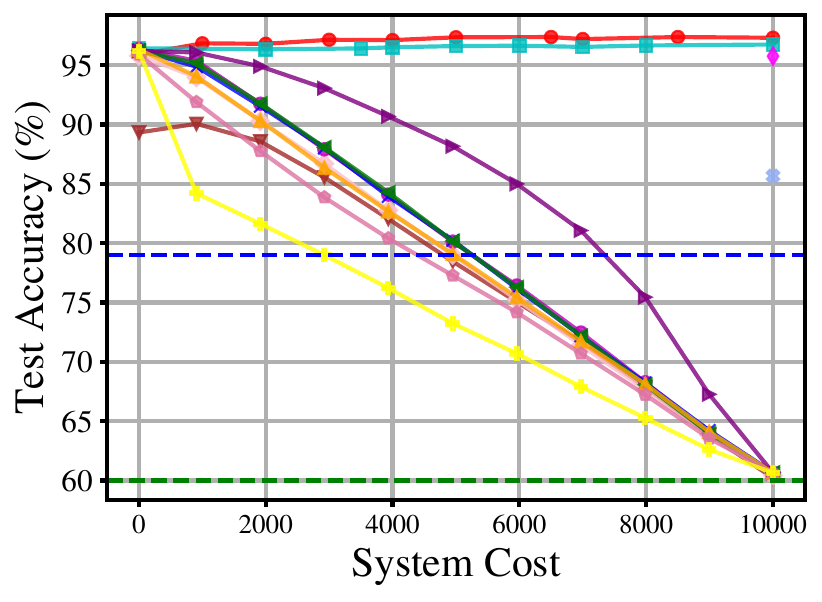}
    \caption{IDN40-aggregation.}
    \label{fig:idn40_aggre_exp}
  \end{subfigure}
  \begin{subfigure}{.32\linewidth}
    \centering
    \includegraphics[width=\linewidth]{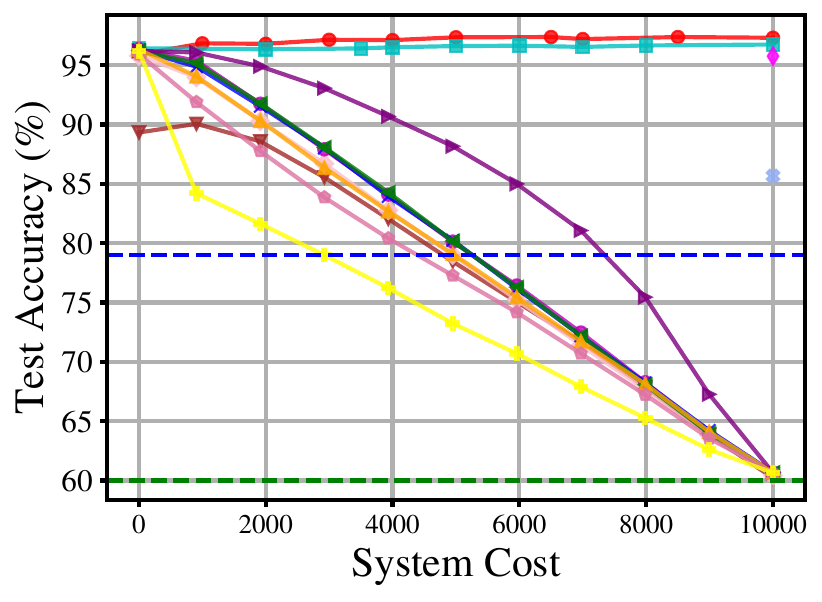}
    \caption{IDN40-random.}
    \label{fig:idn40_random_exp}
  \end{subfigure}
  \begin{subfigure}{.32\linewidth}
    \centering
    \includegraphics[width=\linewidth]{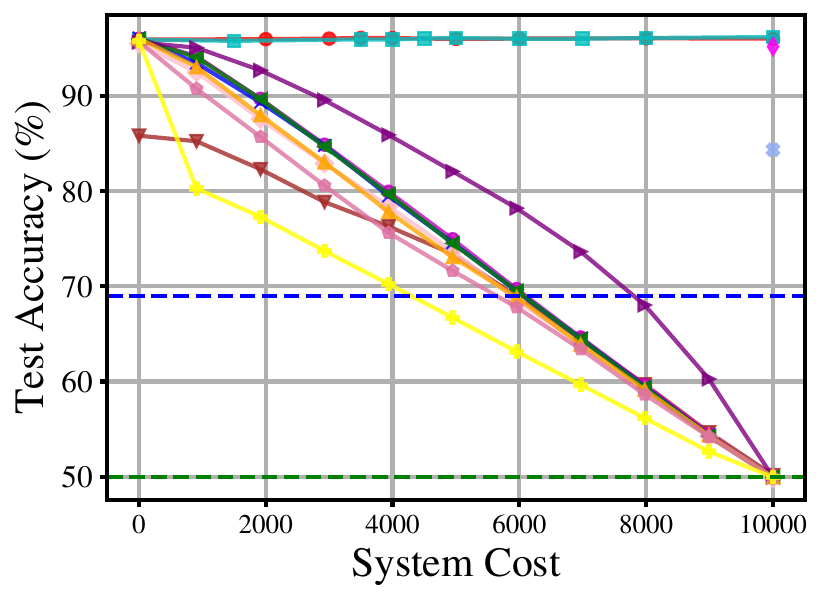}
    \caption{IDN50-aggregation.}
    \label{fig:idn50_aggre_exp}
  \end{subfigure}
  \begin{subfigure}{.32\linewidth}
    \centering
    \includegraphics[width=\linewidth]{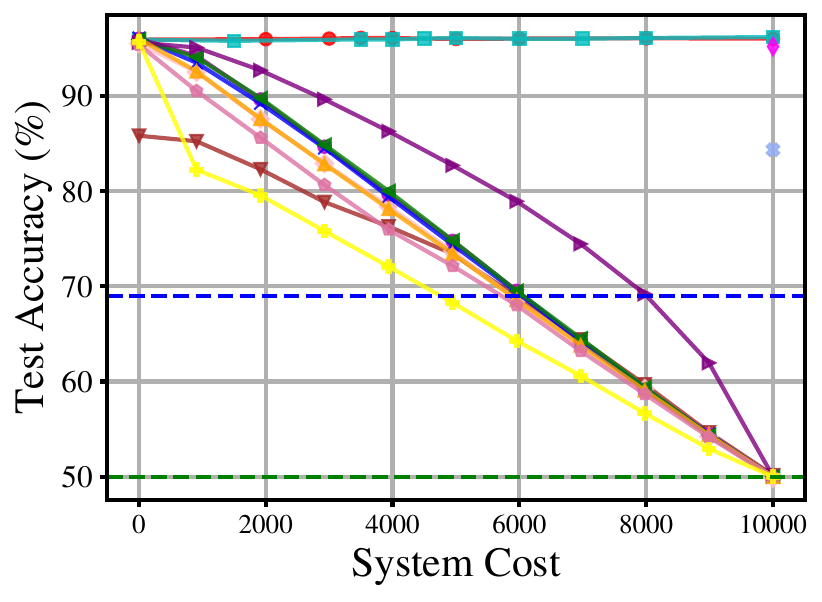}
    \caption{IDN50-random.}
    \label{fig:idn50_random_exp}
  \end{subfigure}
  \begin{subfigure}{.32\linewidth}
    \centering
    \includegraphics[width=\linewidth]{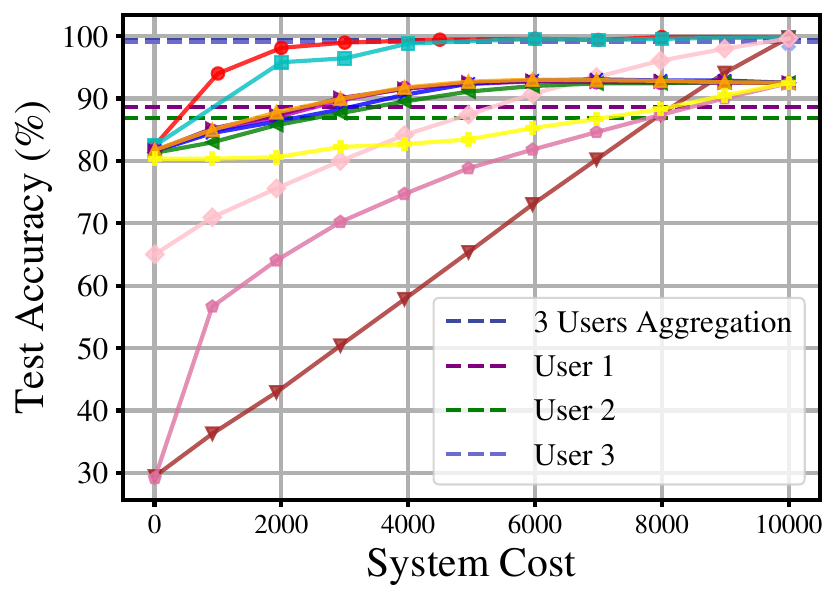}
    \caption{Chaoyang-aggregation.}
    \label{fig:chaoyang_aggre_exp}
  \end{subfigure}
  \begin{subfigure}{.32\linewidth}
    \centering
    \includegraphics[width=\linewidth]{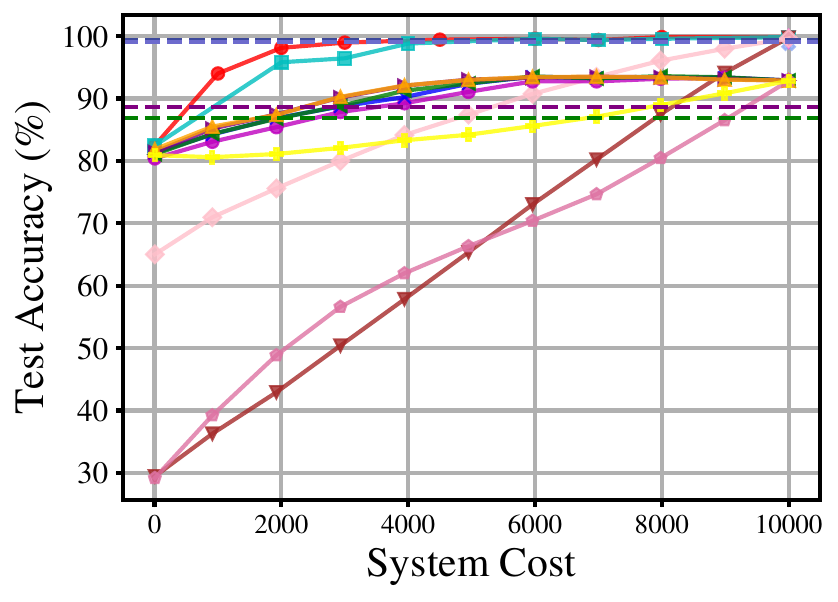}
    \caption{Chaoyang-random.}
    \label{fig:chaoyang_random_exp}
  \end{subfigure}
  \caption{Test accuracy vs. collaboration cost of LECODU (Ours) and competing SEL2D~\cite{whoshould_mozannar23} and ME\{L2D,L2C\}~\cite{ijcai2022-344,multil2d,zhang2023learning} methods. The SEL2D methods are always pre-trained with LNL techniques, with the single user being simulated with either aggregation (majority voting) or random selection (random) from the pool of three annotators. CET and Multi\_L2D show results with and without LNL, while LECOMH shows results with LNL. CET always relies on three experts, resulting in a single point of accuracy vs. cost in each graph. Multi\_L2D can defer to one of many experts, so we select the label corresponding to the maximum probability of 3 users for each sample to draw the curve.
  We truncate the accuracy for all methods at cost=10000. 
  }
  \label{fig:exp}
\end{figure}

\cref{fig:exp} illustrates the cost-accuracy curves for LECODU and other Human-AI Collaborative Classification (HAI-CC) strategies, with detailed numerical results in~\cref{tab:exp} in the supp. material. 
The user accuracy (`User': dashed lines) reflects the inherent noise level of the benchmark, such as around 80\% and 50\% accuracy for IDN20 and IDN50, respectively.
Note that in CIFAR-10H, the LNL AI model (with accuracy $\approx97\%$) surpasses human accuracy ($\approx95\%$), while in Chaoyang, the LNL AI model's accuracy is slightly inferior ($\approx82\%$) compared to that of pathologists (users have accuracy from 86\% to 99\%). Within the IDN benchmarks, the LNL AI consistently outperforms humans.

LECODU shows higher classification accuracy than all competing HAI-CC methods for all collaboration costs in all benchmarks. 
For the majority of competing methods (except LECOMH and CET), the accuracy at cost $=0$ is mostly from the LNL pre-trained model. Then this accuracy increases, reaching a peak for some cost $<10000$, followed by a decrease until reaching the accuracy of users at cost $=10000$. 
On the other hand, MEL2C methods (LECOMH and CET) tend to always improve accuracy with increasing collaboration cost, which is also true for our LECODU.
Note that when the AI model outperforms humans (\eg CIFAR-10H), both our method and competing methods can show more accurate predictions than humans or AI alone. 
In contrast, if humans are more accurate than AI (\eg Chaoyang), the accuracy of SEL2D methods is limited by expert information and cannot improve significantly. However, ME\{L2D,L2C\} methods generally outperform SEL2D approaches as the collaboration cost increases,  
but at the lower cost level, they all fall short of LECODU's accuracy.

In IDN benchmarks, CET (w. LNL) consistently incurs a fixed high cost of $10,000$, showing higher accuracy than SEL2D approaches, but lagging behind LECODU. The performance of Multi\_L2D (w. LNL) aligns closely with that of SEL2D methods. 
Furthermore, LECOMH is often less accurate than LECODU, suggesting that a strategy focusing solely on MEL2C  might fail in some cases where L2D would be a better option. 
Furthermore, LECOMH shows comparable results to LECODU at higher collaboration costs but significantly lags at low-cost levels (e.g., CIFAR-10H, Chaoyang, and IDN20), underlining LECODU's superior adaptability and efficiency in optimising human-AI collaboration for enhanced decision-making accuracy.

\begin{table}[t!]
\centering
\scalebox{0.45}{
\begin{tabular}{lccccc}
\toprule
Image & Human Labels ($\mathcal{M}$) & AI Prediction ($f_{\theta}(.)$) & Selection Probability Vector ($g_{\phi}(.)$) & System Prediction ($h_{\psi}(.)$) & GT \\ \midrule
\begin{minipage}[b]{0.1\columnwidth}
	\centering
	\raisebox{-.5\height}{\includegraphics[width=\linewidth]{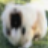}}
\end{minipage} & cat,dog,cat       & dog        & {[}0.91, 0.01, 0.01, 0.03, 0.00, 0.01, 0.03{]} & dog         & dog  \\
\begin{minipage}[b]{0.1\columnwidth}
	\centering
	\raisebox{-.5\height}{\includegraphics[width=\linewidth]{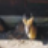}}
\end{minipage} & horse,bird,frog       & deer        & {[}0.95, 0.01, 0.00, 0.02, 0.00, 0.00, 0.02{]} & deer         & deer  \\
\begin{minipage}[b]{0.1\columnwidth}
	\centering
	\raisebox{-.5\height}{\includegraphics[width=\linewidth]{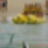}}
\end{minipage} & cat,plane,cat       & ship        & {[}0.92, 0.01, 0.01, 0.03, 0.01, 0.01, 0.03{]} & ship         & ship  \\
\begin{minipage}[b]{0.1\columnwidth}
	\centering
	\raisebox{-.5\height}{\includegraphics[width=\linewidth]{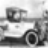}}
\end{minipage} & car,car,car      & truck        & {[}0.93, 0.01, 0.01, 0.02, 0.01, 0.01, 0.01{]} & car         & car  \\
\begin{minipage}[b]{0.1\columnwidth}
	\centering
	\raisebox{-.5\height}{\includegraphics[width=\linewidth]{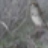}}
\end{minipage} & bird,cat,bird      & frog        & {[}0.90, 0.01, 0.01, 0.03, 0.00, 0.01, 0.04{]} & bird         & bird  \\
\bottomrule 
\end{tabular}}
\caption{LECODU classification (cost $\approx 2000$) of CIFAR-10H test samples, where $\mathcal{M}$ denotes human labels, $f_{\theta}(.)$ is the LNL AI model's classification, $g_{\phi}(.)$ represents the Human-AI Selection prediction probability vector for [AI prediction (1st value), AI + 1 User (2nd value), AI + 2 Users (3rd value), AI + 3 Users (4th value), 1 User (5th value), 2 Users (6th value), 3 Users (7th value)], $h_{\psi}(.)$ is the final prediction from the Collaboration Module, and GT denotes the ground truth label. 
}
\label{tab:case}
\end{table}

In high noise rate scenarios (e.g., IDN\{30,40,50\}), where user annotations are unreliable, the efficacy of HAI-CC reduces, leading to a limited improvement in accuracy by all methods. 
SEL2D methods show a pronounced degradation in performance as collaboration costs increase, where the top accuracy happens at cost $=0$, when the pre-trained LNL method is running alone without any human collaboration. 
For ME\{L2D,L2C\} methods, CET and LECOMH exhibit superior accuracy to SEL2D, but they still fall short of LECODU, while Multi\_L2D also shows similar results as SEL2D methods. 
Although LECODU aligns with LECOMH on IDN50, we can notice a discernible improvement on IDN30 and IDN40, suggesting LECODU can be effective even in highly noisy environments.



\cref{tab:case} shows LECODU's classification results (at cost $\approx 2000$) for CIFAR-10H test samples. The table includes test images, human labels ($\mathcal{M}$), AI predictions (class with maximum probability) from the pre-trained LNL AI model $f_{\theta}(.)$, the prediction probability vector from the Human-AI Selection Module $g_{\phi}(.)$, LECODU's prediction via the Collaboration Module $h_{\psi}(.)$, and the ground truth (GT) label. Notably, when either AI or humans make mistakes, LECODU's prediction tends to be correct, highlighting the system's robustness.

\subsection{Ablation study}
\begin{figure}[t!]
  \centering
    \begin{subfigure}{.24\linewidth}
    \centering
    \includegraphics[width=\linewidth]{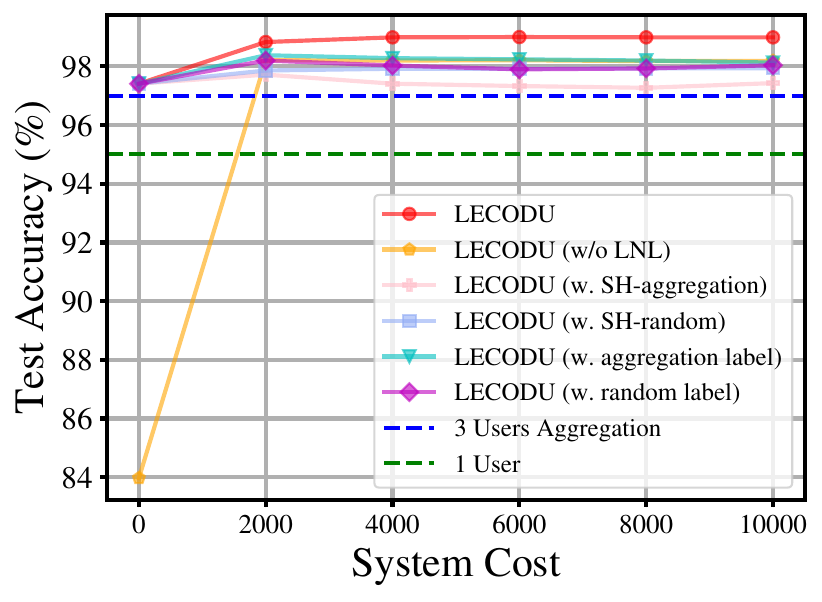}
    \caption{CIFAR-10H.}
    \label{fig:ablation_cifar10h}
  \end{subfigure}
  \begin{subfigure}{.24\linewidth}
    \centering
    \includegraphics[width=\linewidth]{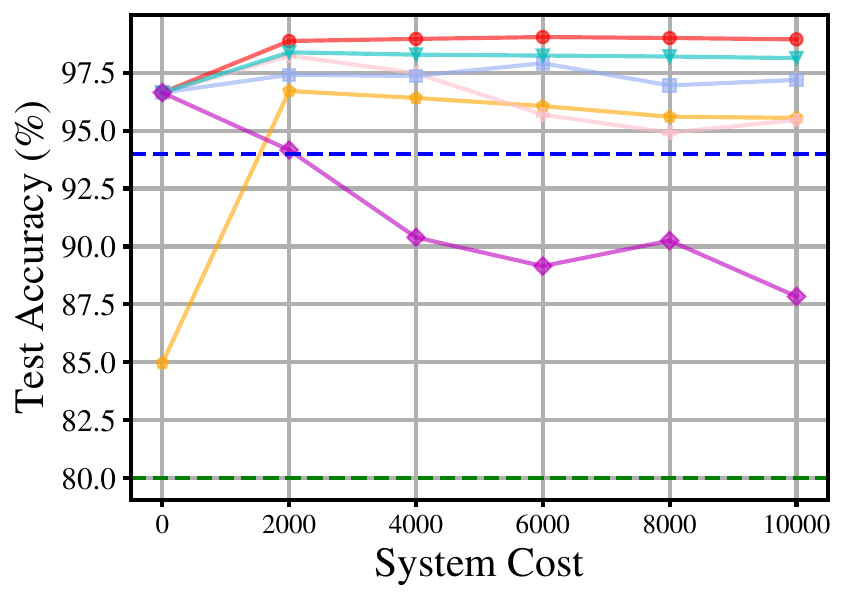}
    \caption{IDN20.}
    \label{fig:ablation_idn20}
  \end{subfigure}
  \begin{subfigure}{.24\linewidth}
    \centering
    \includegraphics[width=\linewidth]{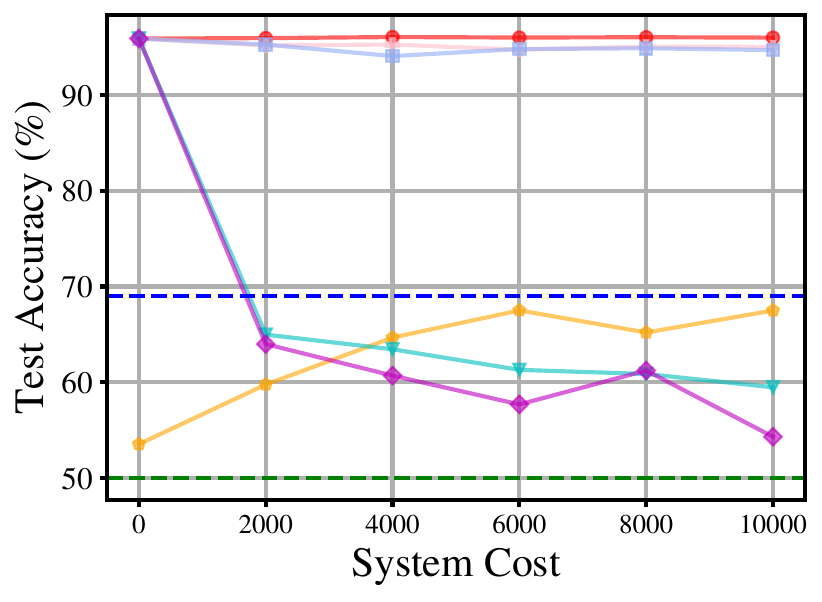}
    \caption{IDN50.}
    \label{fig:ablation_idn50}
  \end{subfigure}
  \begin{subfigure}{.24\linewidth}
    \centering
    \includegraphics[width=\linewidth]{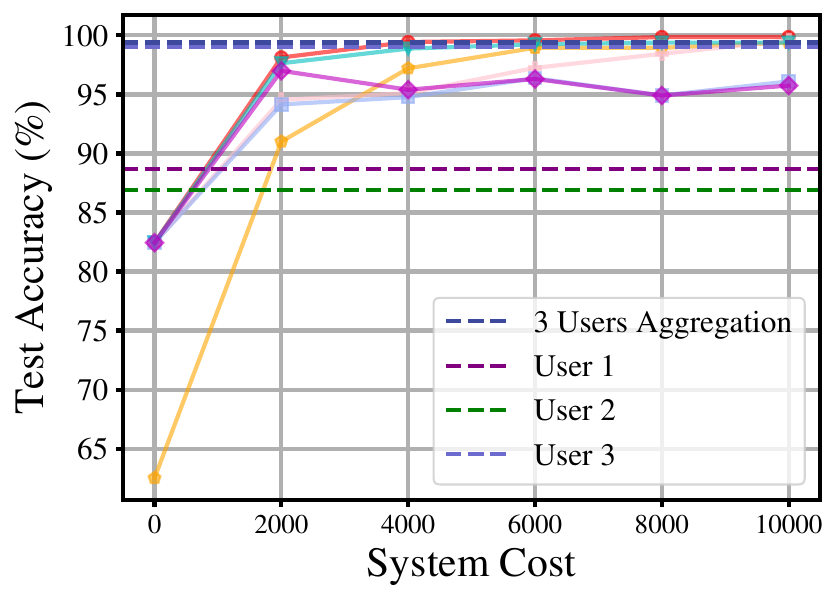}
    \caption{Chaoyang.}
    \label{fig:ablation_chaoyang}
  \end{subfigure}
  
  \caption{Test accuracy vs collaboration cost of LECODU, LECODU without label noise learning (w/o LNL); 
  LECODU (w. SH-aggregation) and (w. SH-random), denoting the reliance on a collaboration with single users (rather than multiple users) formed by aggregating labels and randomly selecting labels, respectively; 
  and
  LECODU (w. aggregation label) and (w. random label), representing LECODU trained with a consensus label using majority vote and a randomly sampled training label, respectively. 
  }
  \label{fig:ablation}
\end{figure}





In~\cref{fig:ablation}, we analyse the influence of some key points of LECODU, namely: 1) engaging multiple users for collaboration during training and testing; 
2) training with CROWDLAB's consensus labels; and
3) pre-training with an LNL AI model~\cite{zhu2021hard,wang2022promix,garg2023instance}.
The comparison between collaborations involving a single user (LECODU w. SH-aggregation or random, where training uses a single label from aggregation through consensus or randomly selected, respectively, and testing relies on a randomly selected label) and multiple users (LECODU) highlights the significance of the latter. 
Multiple users collaboration mitigates the bias inherent in single-user collaborations, a benefit that is particularly pronounced in low noise rate problems (CIFAR-10H, Chaoyang, IDN20), where multi-user approaches consistently surpass single-user approaches at all cost levels, verifying that the lack of multiple users collaboration procedure has a negative impact in such scenarios. 
However, in scenarios with high noise rates, there is little difference between single and multiple users collaborative classification. 

Next, we analyse the differences among MRL approaches to emphasize the importance of generating reliable labels during training. Substituting CROWDLAB's consensus labelling with simpler schemes like majority voting or random labelling (LECODU w. aggregation or random label, respectively) reveals the detrimental effect of relying on unreliable training labels across varying noise levels. 
Such an impact is severe in high noise rates scenarios (e.g., IDN50), but even in lower-noise rates settings (e.g., IDN20, Chaoyang), resorting to random labels significantly degrades model performance.
To highlight even further the significance of CROWDLAB, we evaluate the accuracy of training set consensus labels generated by other MRL methods, such as majority voting and SOTA UnionNet-B~\cite{wei2022deep}, as displayed in \cref{tab:consensus}. Compared to majority voting, UnionNet-B can improve the consensus labels in medium and high noise rate scenarios (e.g., IDN30, IDN40 and IDN50), but lags behind CROWDLAB. The results reflect CROWDLAB's superior capability in producing more reliable and accurate consensus labels. 
\begin{table}[t!]
\centering
\scalebox{0.8}{
\begin{tabular}{lcccccc}
\hline
         & CIFAR-10N & IDN20 & IDN30 & IDN40 & IDN50 & Chaoyang \\ \hline
Majority Vote       & 0.91      & 0.94  & 0.88  & 0.79  & 0.69  & 0.99     \\
UnionNet-B       & 0.92      & 0.94  & 0.93  & 0.93  & 0.90  & 0.99     \\
CROWDLAB & 0.98     & 0.99  & 0.99 & 0.98 & 0.98 & 0.99     \\ \hline
\end{tabular}}
\caption{Accuracy of the consensus label from the majority vote, UnionNet-B and CROWDLAB in the training set.}
\label{tab:consensus}
\end{table}
Furthermore, we conduct experiments without LNL AI model pre-training (LECODU w/o LNL) in Fig.~\ref{fig:ablation}, which confirms that LNL pre-training is crucial, particularly in situations where the LNL AI model's performance exceeds that of the original human annotations (e.g., CIFAR-10H and IDN\{20,50\}). 
The absence of LNL pre-training is detrimental in high-noise environments (e.g., IDN50), as well as in conditions of constrained collaboration budgets  (e.g., Chaoyang, IDN20). Moreover, in scenarios with low-noise rate users (e.g., Chaoyang), the negative impact of removing LNL pre-training becomes pronounced at reduced cost levels.

\begin{figure}[t!]
\centering
\scalebox{0.7}{
  \begin{subfigure}{.49\linewidth}
    \centering
    \includegraphics[width=\linewidth]{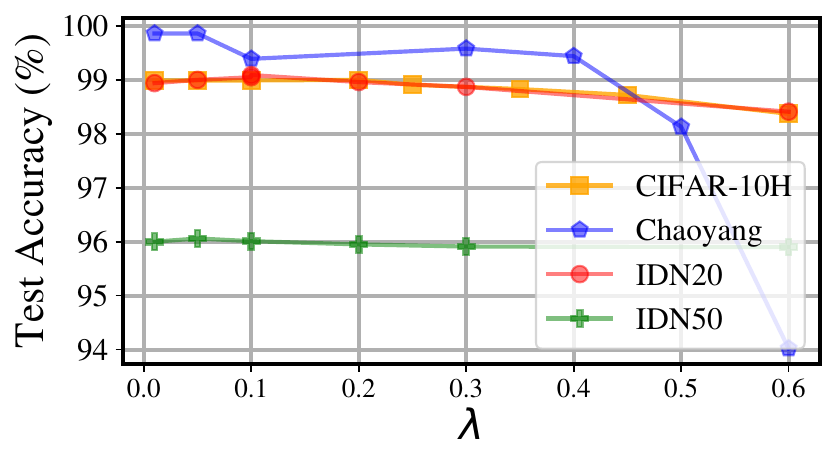}
    \caption{Test accuracy vs $\lambda$.}
    \label{fig:lambda_acc}
  \end{subfigure}
  \begin{subfigure}{.49\linewidth}
    \centering
    \includegraphics[width=\linewidth]{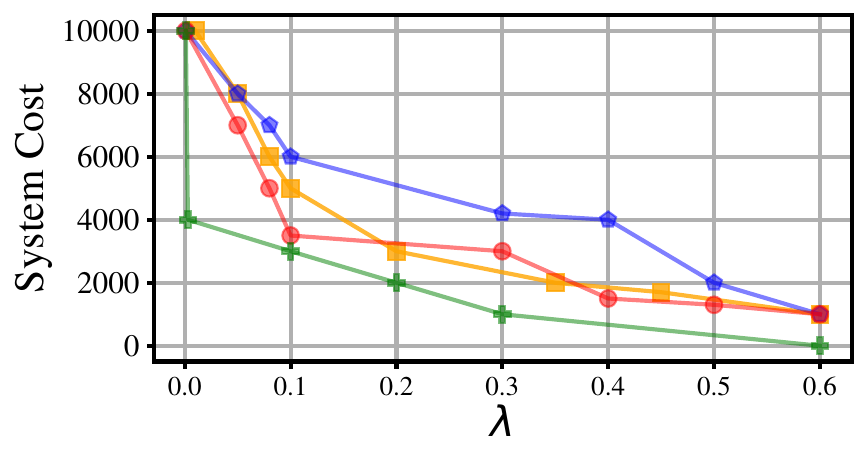}
    \caption{System cost vs $\lambda$.}
    \label{fig:lambda_cost}
  \end{subfigure}}
\caption{Test accuracy and collaboration cost as a function of $\lambda$ in Equation~\ref{eq:loss_function} that weights the collaboration cost in our optimisation.}
\label{fig:lambda}
\end{figure}

In \cref{fig:lambda}, we study the effect of the hyper-parameter $\lambda$, in Eq.~\ref{eq:loss_function}, on the test accuracy and the collaboration cost of the human-AI collaboration system for LECODU on CIFAR-10H, Chaoyang, IDN20, and IDN50. The graphs in this figure display a notable trend where an increase in $\lambda$ corresponds to a decrease in the frequency of human-AI collaborative interactions, thereby reducing overall collaboration costs and accuracy. Interestingly, \cref{fig:exp} shows that within the cost range of [2000, 4000], LECODU reaches its peak performance, 
indicating that the strategy of maximizing expert information integration into the system does not necessarily yield the best outcomes for LECODU.

\begin{figure}[t!]
\centering
\scalebox{0.8}{
\begin{subfigure}{.49\linewidth}
    \centering
    \includegraphics[width=\linewidth]{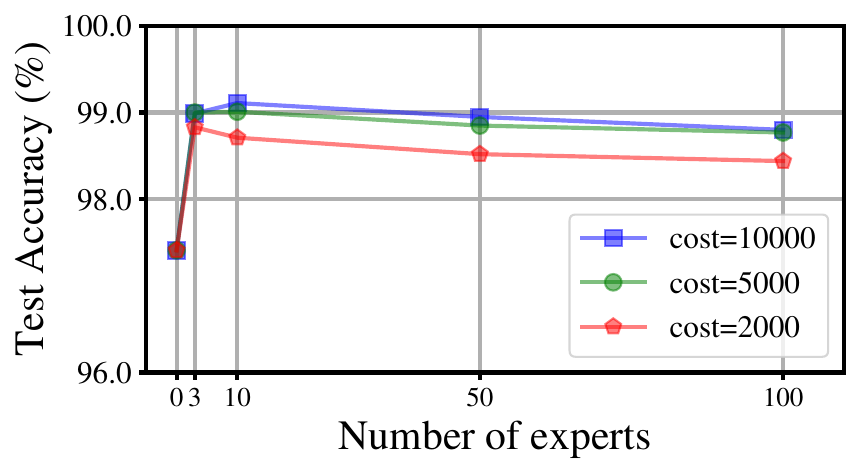}
    \caption{CIFAR-10H.}
    \label{fig:cifar10h_n_usrs}
  \end{subfigure}
  \begin{subfigure}{.49\linewidth}
    \centering
    \includegraphics[width=\linewidth]{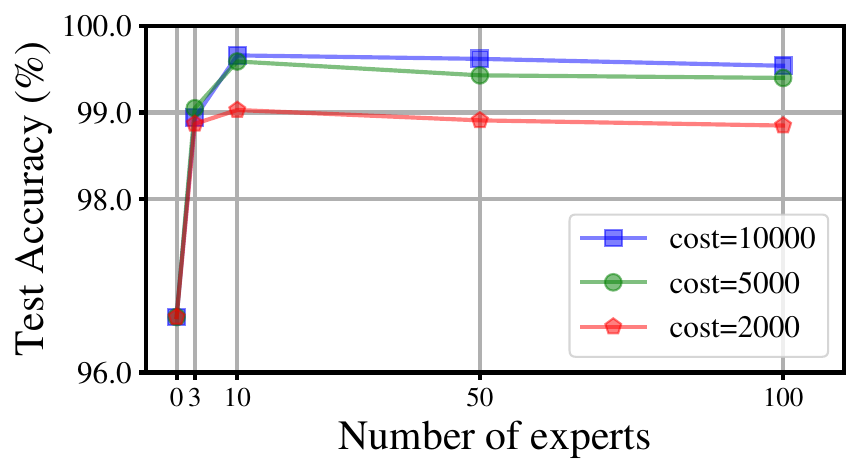}
    \caption{IDN20.}
    \label{fig:idn20_n_usrs}
  \end{subfigure}}
\caption{Test accuracy vs number of experts at different costs. 
}
\label{fig:numofusers}
\end{figure}

We then explore the scalability of LECODU when increasing the number of users to collaborate. \cref{fig:numofusers} shows LECODU's effectiveness on CIFAR-10H and IDN20 across various collaboration cost levels, where the number of available users increases from 0 to 100. 
The run-time complexity of our optimisation has a linear increase in terms of the number of users. In practice, the training time increases from 30s (3 users) to 31s (100 users) per epoch for both CIFAR-10H and IDN20, while the testing time increases from 2s (3 users) to 3s (100 users)  for all 10000 testing images. For CIFAR-10N, the training phase requires simulating numerous users beyond the three available. Following LECOMH~\cite{zhang2023learning}, we synthesise labels from user-specific label-transition matrices of CIFAR-10N's 3 users. Specifically, for each image, we randomly selected one of the label-transition matrices corresponding to the three available users. Following this selection, an annotation for the image was then sampled based on the probability distribution defined by the chosen label-transition matrix. 
For CIFAR-10H and IDN20, accuracy peaks at 3 to 10 users, demonstrating that increasing the number of users required for effective collaboration enables the system to integrate more effective information and improve predictions.

In \cref{tab:traing_time}, we compare the training time per epoch of LECODU and competing methods on CIFAR-10N. 
Notice that the training time of LECODU is competitive to most L2D and L2C methods, but is much shorter than DIFT~\cite{okati2021differentiable}. During the testing phase, the inference time for all methods is about 1s to 2s for all 10000 testing images, not considering the human annotation time.
\begin{table}[]
\centering
\scalebox{1}{
\begin{tabular}{l|c|c|c|c|c|c|c|c|c|c|c|}
\hline
Method & RS  & MoE & LCE & OvA & SP  & CC  & DIFT & CET & Multi\_L2D & LECOMH & LECODU \\ \hline
Time   & 23s & 12s & 24s & 13s & 17s & 24s & 98s  & 32s & 11s        & 30s    & 30s    \\ \hline
\end{tabular}}
\caption{Training time/epoch of LECODU and competing methods on CIFAR-10N.}
\label{tab:traing_time}
\end{table}

\vspace{-40pt}
\section{Conclusion}
\vspace{-5pt}

In this paper, we introduced a novel MEHAI-CC method LECODU. 
LECODU  combines \textit{learning to defer} and \textit{learning to complement} strategies to make three key decisions: when to complement with expert decisions, when to defer decisions to experts, and how many experts should be engaged in the decision process. LECODU not only maximises classification accuracy but also minimises collaboration costs. Comprehensive evaluations across real-world and synthesized multiple noisy label datasets demonstrate LECODU's superior accuracy to SOTA human-AI collaborative classification methods. Even in medium and high noise rate scenarios where user annotations are unreliable, LECODU has a discernible improvement over human decision-makers and AI alone.

One potential weakness of LECODU is that we assume all users are the same, which helps the training process in two ways. First, it mitigates the combinatorial explosion of the number of different combinations of specific users and AI model in L2D and L2C scenarios.
Second, we do not need to identify users, which prevents potential privacy concerns.
However, such assumption is unrealistic since users are different and have distinct classification accuracy that need to be taken into account for training and testing. We plan to propose new AI systems that collaborate  with users with different competencies.
Another issue of LECODU is that it can promote over-reliance on AI in human-AI collaboration, leading to a reduction in the decision-making skills of humans. We plan to introduce methods to mitigate such deskilling problem.  

\bibliographystyle{splncs04}
\bibliography{main}

\clearpage
\setcounter{page}{1}

\section*{Learning to Complement and to Defer to Multiple Users - Supplementary Material}
\subsection*{Quantitative Results}
\cref{tab:exp} shows the test accuracy for our proposed LECODU and competing HAI-CC strategies at different collaboration cost levels in all benchmarks. For SEL2D we select the best results of training with random selected and aggregation labels. For CET and Multi\_L2D, we select the best results of these methods with and without LNL pre-training. The best results of the comparison are given in red, and the second best scores are blue. LECODU shows higher classification accuracy than almost all competing HAI-CC methods in all benchmarks.
\begin{table}[]
\centering
\scalebox{0.83}{
\begin{tabular}{lc|ccccccccccc}
\hline
\multirow{2}{*}{}                               & \multirow{2}{*}{Cost} & \multicolumn{11}{c}{Methods}                                                                 \\ \cline{3-13} 
                                                &                       & RS    & MoE   & LCE   & OvA   & SP    & CC    & DIFT  & CET   & Multi\_L2D & LECOMH & LECODU \\ \hline
\multicolumn{1}{l|}{\multirow{4}{*}{CIFAR-10H}} & 2000                  & 97.68 & 96.84 & 97.82 & 97.76 & 98.24 & 97.55 & 96.82 & -     & 96.51      & \color{blue}98.78  & \color{red}98.83  \\
\multicolumn{1}{l|}{}                           & 4000                  & 96.78 & 96.34 & 97.33 & 97.03 & 97.81 & 97.21 & 96.38 & -     & 96.59      & \color{blue}98.72  & \color{red}99.00  \\
\multicolumn{1}{l|}{}                           & 6000                  & 96.18 & 95.96 & 96.90 & 96.37 & 97.22 & 96.78 & 96.11 & -     & 96.14      & \color{blue}98.77  & \color{red}99.00  \\
\multicolumn{1}{l|}{}                           & 10000                 & 95.17 & 95.17 & 95.17 & 95.17 & 95.17 & 95.17 & 95.17 & 97.76 & 95.46      & \color{blue}98.82  & \color{red}98.99  \\ \hline
\multicolumn{1}{l|}{\multirow{4}{*}{IDN20}}     & 2000                  & 95.85 & 90.10 & 95.83 & 96.06 & 97.53 & 95.43 & 92.56 & -     & 95.08      & \color{blue}98.25  & \color{red}98.87  \\
\multicolumn{1}{l|}{}                           & 4000                  & 91.66 & 87.10 & 91.82 & 91.83 & 97.13 & 91.47 & 88.55 & -     & 91.59      & \color{blue}98.81  & \color{red}99.09  \\
\multicolumn{1}{l|}{}                           & 6000                  & 87.79 & 84.36 & 87.54 & 87.77 & 96.04 & 87.67 & 86.28 & -     & 87.52      & \color{blue}98.60  & \color{red}99.05  \\
\multicolumn{1}{l|}{}                           & 10000                 & 79.85 & 79.84 & 79.85 & 79.85 & 79.85 & 79.85 & 79.85 & 96.13 & 79.61      & \color{blue}98.45  & \color{red}98.94  \\ \hline
\multicolumn{1}{l|}{\multirow{4}{*}{IDN30}}     & 2000                  & 94.00 & 88.8  & 94.01 & 94.31 & 96.97 & 92.92 & 90.45 & -     & 92.62      & \color{blue}96.96  & \color{red}97.42  \\
\multicolumn{1}{l|}{}                           & 4000                  & 88.16 & 83.31 & 88.13 & 88.16 & 95.26 & 86.94 & 84.79 & -     & 87.15      & \color{blue}96.93  & \color{red}97.92  \\
\multicolumn{1}{l|}{}                           & 6000                  & 81.95 & 78.20 & 88.16 & 82.06 & 92.63 & 81.37 & 79.60 & -     & 81.25      & \color{blue}96.96  & \color{red}97.92  \\
\multicolumn{1}{l|}{}                           & 10000                 & 70.33 & 70.34 & 70.34 & 70.34 & 70.34 & 70.34 & 70.34 & 96.52 & 70.39      & \color{blue}97.31  & \color{red}98.20  \\ \hline
\multicolumn{1}{l|}{\multirow{4}{*}{IDN40}}     & 2000                  & 91.75 & 88.17 & 91.54 & 91.81 & 94.90 & 91.12 & 87.99 & -     & 90.29      & \color{blue}96.34  & \color{red}96.80  \\
\multicolumn{1}{l|}{}                           & 4000                  & 84.10 & 80.29 & 83.95 & 84.28 & 90.69 & 83.17 & 80.41 & -     & 82.77      & \color{blue}96.51  & \color{red}97.14  \\
\multicolumn{1}{l|}{}                           & 6000                  & 76.43 & 72.48 & 76.18 & 76.29 & 85.00 & 75.45 & 74.17 & -     & 75.25      & \color{blue}96.64  & \color{red}97.37  \\
\multicolumn{1}{l|}{}                           & 10000                 & 60.67 & 60.67 & 60.67 & 60.67 & 60.67 & 60.67 & 60.67 & 95.76 & 60.41      & \color{blue}96.74  & \color{red}97.31  \\ \hline
\multicolumn{1}{l|}{\multirow{4}{*}{IDN50}}     & 2000                  & 89.69 & 79.55 & 89.21 & 89.75 & 92.68 & 87.59 & 85.6  & -     & 87.55      & \color{blue}95.77  & \color{red}95.95  \\
\multicolumn{1}{l|}{}                           & 4000                  & 79.64 & 72.07 & 79.41 & 80.00 & 86.28 & 78.13 & 75.95 & -     & 78.40       & \color{blue}95.91  & \color{red}96.06  \\
\multicolumn{1}{l|}{}                           & 6000                  & 69.53 & 64.27 & 69.39 & 69.53 & 78.92 & 68.58 & 68.02 & -     & 68.94      & \color{blue}96.07  & \color{red}96.07  \\
\multicolumn{1}{l|}{}                           & 10000                 & 50.03 & 50.03 & 50.03 & 50.03 & 50.03 & 50.03 & 50.03 & 95.18 & 50.17      & \color{red}96.17  & \color{blue}96.12  \\ \hline
\multicolumn{1}{l|}{\multirow{4}{*}{Chaoyang}}  & 2000                  & 87.00 & 81.11 & 86.86 & 86.76 & 87.47 & 87.79 & 64.04 & -     & 75.60      & \color{blue}95.82  & \color{red}98.13  \\
\multicolumn{1}{l|}{}                           & 4000                  & 91.67 & 83.31 & 90.60 & 91.25 & 92.05 & 92.09 & 74.71 & -     & 84.15      & \color{blue}98.82  & \color{red}99.44  \\
\multicolumn{1}{l|}{}                           & 6000                  & 92.75 & 85.60 & 93.50 & 93.55 & 93.31 & 93.45 & 81.81 & -     & 90.74      & \color{blue}99.58  & \color{red}99.86  \\
\multicolumn{1}{l|}{}                           & 10000                 & 92.89 & 92.89 & 92.89 & 92.89 & 92.89 & 92.89 & 92.89 & 99.58 & 99.76      & \color{blue}99.86  & \color{red}99.86  \\ \hline
\end{tabular}}
\caption{
Test accuracy (\%) vs. collaboration cost of LECODU (Ours) and competing SEL2D~\cite{whoshould_mozannar23} and ME\{L2D,L2C\}~\cite{ijcai2022-344,multil2d,zhang2023learning} methods. The SEL2D methods are always pre-trained with LNL techniques, with the single user being simulated with either aggregation (majority voting) or random selection (random) from the pool of three annotators. Here we show the best results of them. CET, Multi\_L2D show the best results with and without LNL, while LECOMH show results with LNL. CET always relies on three experts, resulting in a single value at the maximum of cost. Multi\_L2D can defer to one of many experts, so we select the label corresponding to the maximum probability of 3 users for each sample. We truncate the accuracy for all methods at cost=10000. The best results are given in red, and the second best scores are blue.}
\label{tab:exp}
\end{table}
\end{document}